\documentclass[prodmode,acmtoit]{acmsmall} 

\usepackage[ruled]{algorithm2e}
\usepackage{amsmath}
\usepackage{hyperref}
\usepackage{color}
\usepackage{ulem}
\usepackage[utf8]{inputenc}

\usepackage{lastpage}
\usepackage{fancyhdr}

\renewcommand{\algorithmcfname}{ALGORITHM}
\SetAlFnt{\small}
\SetAlCapFnt{\small}
\SetAlCapNameFnt{\small}
\SetAlCapHSkip{0pt}
\IncMargin{-\parindent}

\begin{document}

\markboth{L. Sun et al.}{Supervised Anomaly Detection in Uncertain Pseudo-Periodic Data Streams}

\title{Supervised Anomaly Detection in Uncertain Pseudo-Periodic Data Streams}
\author{JIANGANG MA
\affil{Victoria University}
LE SUN
\affil{Victoria University}
HUA WANG
\affil{Victoria University}
YANCHUN ZHANG
\affil{Victoria University}
UWE AICKELIN 
\affil{The University of Nottingham}}

\begin{abstract}
Uncertain data streams have been widely generated in many Web applications. The uncertainty in data streams makes anomaly detection from sensor data streams far more challenging. In this paper, we present a novel framework that supports anomaly detection in uncertain data streams. The proposed framework adopts an efficient uncertainty pre-processing procedure to identify and eliminate uncertainties in data streams. Based on the corrected data streams, we develop effective period pattern recognition and feature extraction techniques to improve the computational efficiency. We use classification methods for anomaly detection in the corrected data stream. We also empirically show that the proposed approach shows a high accuracy of anomaly detection on a number of real datasets.
\end{abstract}





\begin{bottomstuff}
Correspondence author's addresses: Yanchun Zhang, School of Computer Science, Fudan University, Shanghai 200433, China; and
Centre for Applied Informatics,
Victoria University, VIC, 3012, Australia\\
\end{bottomstuff}

\maketitle

\pagestyle{empty}
\section{Introduction}
Data streams have been widely generated in many Web applications such as monitoring click streams \cite{gunduz2003web}, stock tickers \cite{chen2000niagaracq,zhu2002statstream}, sensor data streams and auction bidding patterns \cite{arasu2004cql}. For example, in the applications of Web tracking and personalization, Web log entries or click-streams are typical data streams. Other traditional and emerging applications include wireless sensor networks (WSN) in which data streams collected from sensor networks are being posted directly to the Web. Typical applications comprise environment monitoring (with static sensor nodes) \cite{Akyildiz2005257} and animal and object behaviour monitoring (with mobile sensor nodes), such as water pollution detection \cite{He201269} based on water sensor data, agricultural management and cattle moving habits \cite{CSIRO2011}, and analysis of trajectories of animals \cite{Gudmundsson2007}, vehicles \cite{zheng2010geolife} and fleets \cite{Lee2007}.

Anomaly detection is a typical example of a data streams application. Here, anomalies or outliers or exceptions often refer to the patterns in data streams that deviate expected normal behaviours. Thus, anomaly detection is a dynamic process of finding abnormal behaviours from given data streams. For example, in medical monitoring applications, a human electrocardiogram (ECG) (vital signs) and other treatments and measurements are typical data streams that appear in a form of periodic patterns. That is, the data present a repetitive pattern within a certain time interval. Such data streams are called pseudo periodic time series. In such applications, data arrives continuously and anomaly detection must detect suspicious behaviours from the streams such as abnormal ECG values, abnormal shapes or exceptional period changes.

Uncertainty in data streams makes the anomaly detection far more challenging than detecting anomalies from deterministic data. For example, uncertainties may result from missing points from a data stream, missing stream pieces, or measurement errors due to different reasons such as sensor failures and measurement errors from different types of sensor devices. This uncertainty may cause serious problems in data stream mining. For example, in an ECG data stream, if a sensor error is classified as abnormal heart beat signals, it may cause a serious misdiagnosis. Therefore, it is necessary to develop effective methods to distinguish uncertainties and anomalies, remove uncertainties, and finally find accurate anomalies.

There are a number of related research areas to sensor data stream mining, such as data streams compression, similarity measurement, indexing and querying mechanisms \cite{esling2012time}. For example, to clean and remove uncertainty from data, a method for compressing data streams was presented in \cite{douglas1973algorithms}. This method uses some critical points in a data stream to represent the original stream. However, this method cannot compress uncertain data streams efficiently because such compression may result in an incorrect data stream approximation and it may remove useful information that can correct the error data.

This paper focuses on anomaly detection in uncertain pseudo periodic time series. A pseudo periodic time series refers to a time-indexed data stream in which the data present a repetitive pattern within a certain time interval. However, the data may in fact show small changes between different time intervals. Although much work has been devoted to the analysis of pseudo periodic time series \cite{keogh2005hot,huang2013online}, few of them focus on the identification and correction of uncertainties in this kind of data stream.

In order to deal with the issue of anomaly detection in uncertain data streams, we propose a supervised classification framework for detecting anomalies in uncertain pseudo periodic time series, which comprises four components: a uncertainty identification and correction component (UICC), a time series compression component (TSCC), a period segmentation and summarization component (PSSC), and a classification and anomaly detection component (CADC). First, UICC processes a time series to remove uncertainties from the time series. Then TSCC compresses the processed raw time series to an approximate time series. Afterwards the PSSC identifies the periodic patterns of the time series and extracts the most important features of each period, and finally the CADC detects anomalies based on the selected features. Our work has made the following distinctive contributions:

\begin{itemize}
\item We present a classification-based framework for anomaly detection in uncertain pseudo periodic time series, together with a novel set of techniques for segmenting and extracting the main features of a time series. The procedure of pre-processing uncertainties can reduce the noise of anomalies and improve the accuracy of anomaly detection. The time series segmentation and feature extraction techniques can improve the performance and time efficiency of classification.
\item We propose the novel concept of a feature vector to capture the features of the turning points in a time series, and introduce a silhouette value based approach to identify the periodic points that can effectively segment the time series into a set of consecutive periods with similar patterns.
\item We conduct an extensive experimental evaluation over a set of real time series data sets. Our experimental results show that the techniques we have developed outperform previous approaches in terms of accuracy of anomaly detection. In the experiment part of this paper, we evaluate the proposed anomaly detection framework on ECG time series. However, due to the generic nature of features of pseudo periodic time series (e.g. similar shapes and intervals occur in a periodic manner), we believe that the proposed method can be widely applied to periodic time series mining in different areas.
\end{itemize}

The structure of this paper is as follows: Section 2 introduces the related research work. Section 3 presents the problem definition and generally describes the proposed anomaly detection framework. Section 4 describes the anomaly detection framework in detail. Section 5 presents the experimental design and discusses the results. Finally, Section 6 concludes this paper.

\section{Related Work}

We analyse the related research work from two dimensions: anomaly detection and uncertainty processing.

\textbf{Anomaly detection in data streams:} Anomaly detection in time series has various applications in wide area, such as intrusion detection \cite{Tavallaee2010}, disease detection in medical sensor streams \cite{manning2010array}, and biosurveillance \cite{shmueli2010statistical}. Zhang et al.\cite{Zhang2009581} designed a Bayesian classifier model for identification of cerebral palsy by mining gait sensor data (stride length and cadence). In stock price time series, anomalies exist in a form of change points that reflect the abnormal behaviors in the stock market and often repeating motifs are of interest \cite{wilson2008motif}. Detecting change points has significant implications for conducting intelligent trading \cite{Jiang2011Anomaly}. Liu et al. \cite{Liu2010distribution} proposed an incremental algorithm that detects changes in streams of stock order numbers, in which a Poisson distribution is adopted to model the stock orders, and a maximum likelihood (ML) method is used to detect the distribution changes.

The segmentation of a time series refers to the approximation of the time series, which aims to reduce the time series dimensions while keeping its representative features \cite{esling2012time}. One of the most popular segmentation techniques is the Piecewise Linear Approximation (PLA) based approach \cite{keogh03segmenting,Qi2015TSseg}, which splits a time series into segments and uses polynomial models to represent the segments. Xu et al. \cite{XuTSSeg2012} improved the traditional PLA based techniques by guaranteeing an error bound on each data point to maximally compact time series. Daniel \cite{Daniel2006Piecewise} introduced an adaptive time series summarization method that models each segment with various polynomial degrees. To emphasize the significance of the newer information in a time series, Palpanas et al. \cite{Palpanas2008Amnesic} defined user-oriented amnesic functions for decreasing the confidence of older information continuously.

However, the approaches mentioned above are not designed to process and adapt to the area of pseudo periodic data streams. Detecting anomalies from periodic data streams has received considerable attention and several techniques have been proposed recently \cite{folarin2001holter,grinsted2004application,levy2007cardiovascular}. The existing techniques for anomaly detection adopt sliding windows \cite{keogh2005hot,gu2005detecting} to divide a time series into a set of equal-sized sub-sequences. However, this type of method may be vulnerable to tiny difference in time series because it cannot well distinguish the abnormal period and a normal period having small noisy data. In addition, as the length of periods is varying, it is difficult to capture the periodicity by using a fixed-size window \cite{tang2007effective}. Other examples of segmenting pseudo periods include an peak-point-based clustering method and valley-point-based method \cite{huang2013online,tang2007effective}. These two methods may have very low accuracy when the processed time series have noisy peak points or have irregularly changed sub-sequences. Our proposed approach falls into the category of classification-based anomaly detection, which is proposed to overcome the challenge of anomaly detection in periodic data streams. In addition, our method is able to identify qualified segmentation and assign annotation to each segment to effectively support the anomaly detection in a pseudo periodic data streams.

\textbf{Uncertainty processing in data streams:} Most data streams coming from real-world sensor monitoring are inherently noisy and uncertainties. A lot of work has concentrated on the modelling of uncertain data streams \cite{Aggarwal2008,Aggarwal2009,Leung2009}. Dallachiesa et al.\cite{Dallachiesa2012} surveyed recent similarity measurement techniques of uncertain time series, and categorized these techniques into two groups: probability density function based methods \cite{Sarangi2010} and repeated measurement methods \cite{Johannes2009}. Tran et al.\cite{Tran2012} focused on the problem of relational query processing on uncertain data streams. However, previous work rarely focused on the detection and correction of the missing critical points for a discrete time series. In this work, we model a continuous time series as a discrete time series by identifying the critical points in a time series, and introduce a novel method of detecting and correcting the missing inflexions based on the angles between points.

\section{Problem Specification and Framework Description}
In this section, we first give a formal definition of the problems and then describe the proposed framework of detecting abnormal signals in uncertain time series with pseudo periodic patterns. The symbols frequently used in this paper are summarized in Table \ref{tab:Symbols}.

\begin{table}
\tbl{Frequently Used Symbols\label{tab:Symbols}}{
\begin{tabular}{l*{2}{l}r}
\hline
Symbols & Meaning \\
\hline
$TS$ & A time series \\
$p_{i}$ & The $i$th point in a $TS$ \\
$SS$ & A subsequence \\
$PTS$ & A pseudo periodic time series \\
$Q$ & A set of period points in a $PTS$ \\
$pd$ & A period in a $PTS$ \\
$CTS$ & A compressed $PTS$ \\
$diff_{i}$ & $diff_{1i} = t_{i} - t_{i-1}$, $diff_{2i} = t_{i+1} - t_{i}$ \\
$vec_{i}$ & A feature vector of point $p_{i}$ \\
$sil(p_{i})$ & Silhouette value of point $p_{i}$ \\
$sim(p_{i},p_{j})$ & Euclidean distance based similarity between points $p_{i}$ and $p_{j}$ \\
$C$ & A set of clusters \\
$msil(C)$ & Mean silhouette value of a cluster $C$ \\
$seg_{i}$ & A summary of a period \\
$STS$ & A segmented $CTS$ \\
$A_{STS}$ & A set of annotations \\
$Lbs$ & A set of labels indicating the states \\
$lb_{(i)}$ & The $i$th label in $Lbs_{PTS}$ \\
\hline
\end{tabular}}
\end{table}

\subsection{Problem definition}
\begin{definition}
A \textbf{time-series} $TS$ is an ordered real sequence: $TS=(v_{1},\cdots,v_{n})$, where $v_{i}$, $i \in [1,n]$, is a point value on the time series at time $t_{i}$.
\end{definition}

We use the form $|TS|$ to represent the number of points in time series $TS$ (i.e., $|TS|=n$). Based on the above definition, we define subsequence of a $TS$ as below.

\begin{definition}
For time series $TS$, if $SS(\subset TS)$ comprises $m$ consecutive points: $SS=(v_{s_{1}},\cdots,v_{s_{m}})$, we say that $SS$ is a \textbf{subsequence} of $TS$ with length $m$, represented as $SS\sqsubseteq TS$.
\end{definition}

\begin{definition}
A \textbf{pseudo periodic time series} $PTS$ is a time series $PTS = (v_{1},v_{2},\cdots,v_{n})$, $\exists Q = \{v_{p_{1}},\cdots,v_{p_{k}}|v_{p_{i}}\in PTS, i \in [1,k]\}$, that regularly separates $PTS$ on the condition that
\begin{enumerate}
\item $\forall i\in [1,k-2]$, if $\bigtriangleup_{1}=|p_{i+1}-p_{i}|,\bigtriangleup_{2}=|p_{i+2}-p_{i+1}|$, then $|\bigtriangleup_{2} - \bigtriangleup_{1}|\leq \xi_{1}$; where $\xi_{1}$ is a small value.
\item let $s_{1} = (v_{p_{i}},v_{(p_{i})+1},\cdots,v_{p_{i+1}})\sqsubseteq PTS$, and $s_{2}=(v_{p_{i+1}},v_{(p_{i+1})+1},\cdots,v_{p_{i+2}})\sqsubseteq PTS$, then $dsim(s1,s2) \leq \xi_{2}$, where $dsim()$ calculates the dis-similarity between $s1$ and $s2$, and $\xi_{2}$ is a small value. $dsim()$ can be any dis-similarity measuring function between time series, e.g., Euclidean distance.
\end{enumerate}
In particular, $v_{p_{i+1}}\in Q$ is called a period point.
\end{definition}

An uncertain $PTS$ is a $PTS$ having error detected data or missing points.

\begin{definition}
If $pd \sqsubseteq PTS$, and $pd = (v_{p_{i}},v_{(p_{i})+1},\cdots,v_{p_{i+1}}), \forall v_{p_{i}}\in Q$, then $pd$ is called a \textbf{period} of the $PTS$.
\end{definition}

\begin{definition}
A \textbf{normal pattern} $M$ of a $PTS$ is a model that uses a set of rules to describe a behaviour of a subsequence $SS$, where $m=|SS|$ and $m\in [1,|PTS|/2]$. This behaviour indicates the normal situation of an event.
\end{definition}

Based on the above definitions, we describe types of anomalies that may occur in a $PTS$. There are two possible types of anomalies in a $PTS$: local anomalies and global anomalies Given the $PTS$ in Definition 3.3, and a normal pattern $N=(v_{1},\cdots,v_{m}) \sqsubseteq PTS$, a local anomaly $(L)$ is defined as:

\begin{definition}
Assume $L = (v_{l_{1}},\cdots,v_{l_{n}}) \sqsubseteq PTS$, $L$ is a local anomaly if either of the two conditions in Definition 3.3 is broken (shown as below (1)), and at the same time satisfies the following two conditions (below (3)):
\begin{enumerate}
\item $\bigtriangleup_{N}-\bigtriangleup_{L}>\xi_{1}$ or $dsim(N,L)> \xi_{2}$;
\item frequency of $L$: $freq(L) \ll freq(N)$ and $L$ does not happen in a regular sampling frequency.
\item $|L| \ll |PTS|$.
\end{enumerate}
\end{definition}

\begin{figure}
\centering
\includegraphics[width=8cm, height=5cm]{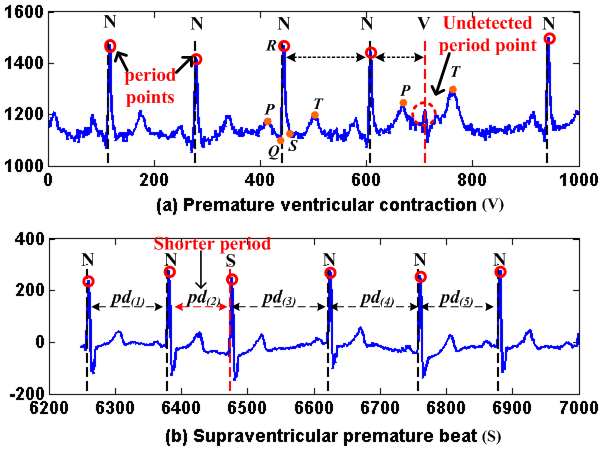}
\caption{Two examples of local anomaly in ECG time series}
\label{fig:localAnomaly}
\end{figure}

\begin{example}
Fig.\ref{fig:localAnomaly} shows two examples of pseudo periodic time series and their local anomalies. Fig.\ref{fig:localAnomaly}(a) shows a premature ventricular contraction signal in an ECG stream. A premature ventricular contraction (PVC) \cite{levy2007cardiovascular} is perceived as a "skipped beat". It can be easily distinguished from a normal heart beat when detected by the electrocardiogram. From Fig.\ref{fig:localAnomaly}(a), the QRS and T waves of a PVC (indicated by V) are very different from the normal QRS and T (indicated by N). Fig.\ref{fig:localAnomaly}(b) presents an example of premature atrial contractions (PACs)\cite{folarin2001holter}. A PAC is a premature heart beat that occurs earlier than the regular beat. If we use the highest peak points as the period points, then a segment between two peak points is a period. From Fig.\ref{fig:localAnomaly}, the second period (a PAC) is clearly shorter than the other periods.
\end{example}

\subsection{Overview of the Anomaly Detection Framework for Uncertain Time Series Data}
\begin{figure}
\centering
\includegraphics[width=10cm, height=2cm]{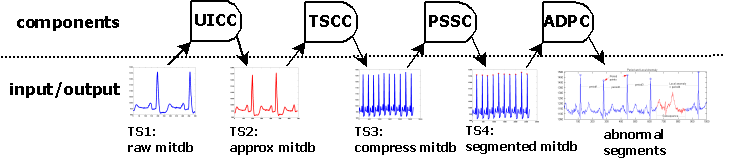}
\caption{Workflow of the $mitdb$ processing based on the proposed framework}
\label{fig:flowchart}
\end{figure}

As mentioned previously, the proposed framework comprises four main components: an uncertainty identification and correction component (UICC), a time series compression component (TSCC), a period segmentation and summarization component (PSSC), and an anomaly detection and prediction component (ADPC). We explain the process of anomaly detection of the proposed framework using an example of the dataset $mitdb$. Fig.\ref{fig:flowchart} shows the processing progress of $mitdb$. First, the raw $mitdb$ time series is an input to the UICC component. The TS$1$ in Fig.\ref{fig:flowchart} shows a subsequence of the raw $mitdb$. The UICC identifies the inflexions (including missing inflexions) of $mitdb$, and the raw $mitdb$ is transformed into an approximated time series that only consists of the identified inflexions (TS$2$ in Fig.\ref{fig:flowchart}). The TSCC component then further compresses the approximated $mitdb$. The TS$3$ in Fig.\ref{fig:flowchart} shows the compressed time series ($CTS$) that is a compression of the subsequence in TS$2$. The PSSC component segments the time series and assigns annotations to each segment. TS$4$ in Fig.\ref{fig:flowchart} shows the segmented and annotated $CTS$ corresponding to the $CTS$ in TS$3$. Finally, the ADPC component learns a classification model based on the segmented $CTS$ to detect abnormal subsequences in similar time series.

In the next section, we introduce the framework and its four components in detail.

\section{Anomaly Detection in Uncertain Periodic Time Series}

\subsection{Uncertainty Identification and Correction: UICC}

In this section, we introduce the procedure of eliminating uncertainties of a $PTS$ caused by non-captured key-points of a $PTS$, based on our previous work \cite{he2013circe}. We first introduce the definition of key-points of a time series.

\begin{definition}
Given a $PTS = (v_{1},\cdots,v_{n})$, if a point, $p_{i} = v_{1}$ or $v_{n}$, is a turning point, then $p_{i}$ is a key-point; or else, if $\angle p_{i} = \pi - \angle p_{j}p_{i}p_{k}$ and $\angle p_{i} > \epsilon$, where $\angle p_{i}$ is the angle between vectors $\overrightarrow{p_{j}p_{i}}$ and $\overrightarrow{p_{i}p_{k}}$, $2 \leq j < i < k \leq n$, $\epsilon$ is a threshold, $p_{j}$ and $p_{k}$ are key-points, and for any point $p_{r}, j < r < k$, $\angle p_{r} \leq \epsilon$, then $p_{i}$ is a key-point.
\end{definition}

\begin{figure}
\centering
\includegraphics[width=9cm, height=2cm]{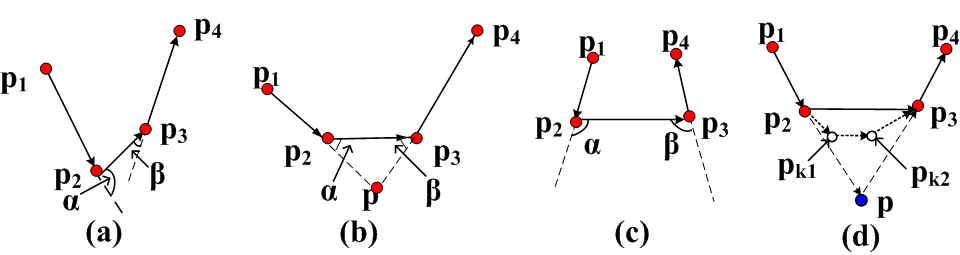}
\caption{(a) $p_{2}$ is a key-point. (b) $p$ is a missing key-point. (c) $p_{2}$ and $p_{3}$ are two key-points. (d) $p_{k1}$ and $p_{k2}$ are two missing key-points, while $p$ is one deduced key-point.}
\label{fig:key-point}
\end{figure}

From the above definition, the core procedure to determine a point $p_{k}$ as a key-point is based on the angles between $\overrightarrow{p_{k-1}p_{k}}$ and $\overrightarrow{p_{k}p_{k+1}} (i.e., \angle p_{k} = \pi - \angle p_{k-1}p_{k}p_{k+1})$, given that $p_{k-1}$ and $p_{k+1}$ are both key-points. If $\angle p_{k}$ is larger than a threshold value, and the angles of all the other points between $k-1$ and $k+1$ are not larger than the threshold, then $p_{k}$ is a key-point. However, if $p_{k}$ is missing, we need to check at least four points: two key-points before and two key-points after $p_{k}$ respectively. Therefore, we generally check four consecutive points at the same time. Combined with Fig.\ref{fig:key-point}, the detailed process is described below:

Given four consecutive points $p_{1} = v_{1}$, $p_{2} = v_{2}$, $p_{3} = v_{3}$, and $p_{4} = v_{4}$, where $p_{1}$ and $p_{4}$ are key-points, and a small value $\epsilon\rightarrow 0$, let $\angle p_{2} = \pi - \angle p_{1}p_{2}p_{3}$ and $\angle p_{3} = \pi - \angle p_{2}p_{3}p_{4}$,
\begin{itemize}

\item If $\angle p_{2} > \epsilon$, $\angle p_{3} < \epsilon$, and there is no other point between $p_{1}$ and $p_{4}$, then $p_{2}$ is a key-point (see Fig.\ref{fig:key-point}(a));

\item If $\angle p_{2} < \epsilon$, and $\angle p_{3} > \epsilon$, then $p_{3}$ is a key-point;

\item If $\angle p_{2} > \epsilon$, $\angle p_{3} > \epsilon$, and $\angle p_{2} + \angle p_{3} < \pi$, then there may be a missing key-point. In this case, it is also possible that both of $p_{2}$ and $p_{3}$ are key-points. If we can find a missing point $p = v$ at time $t$, that $\angle p = \angle p_{2} + \angle p_{3} \geq 2*\epsilon$, then the point $p$ is more likely to be a key-point between $p_{2}$ and $p_{3}$, as the larger $\angle p$ indicates the larger turning degree of the time series at point $p$.
 We deduce missing key-points by solving the equation $Q = \frac{|p_{2}p|}{sin(\angle p_{3})} = \frac{|p_{3}p|}{sin(\angle p_{2})}$, where $Q = \frac{|p_{2}p_{3}|}{sin(\pi - \angle p_{2} - \angle p_{3})}$, which can be written as:

\begin{equation}\label{eq:dedMisInf}
 \left\{
 \begin{array}{c}
 Q^{2}sin^{2}\angle p_{3} = (v - v_{2})^{2} + (t - t_{2})^{2}\\
 Q^{2}sin^{2}\angle p_{2} = (v - v_{3})^{2} + (t - t_{3})^{2}\\
 \end{array}
 \right.
\end{equation}

If Equation \eqref{eq:dedMisInf} only has one solution, this solution is a key-point; if it has two solutions, we adopt the one on the line of $\overline{p_{1}p_{2}}$, i.e., $p$ in Fig.\ref{fig:key-point}(b) as a key-point; if it does not have solution, point $p_{2}$ and $p_{3}$ are key-points.

\item If $\angle p_{2} > \epsilon$, $\angle p_{3} > \epsilon$, and $\angle p_{2} + \angle p_{3} > \pi$, then $p_{2}$ and $p_{3}$ are both key-points (Fig.\ref{fig:key-point}(c)). In addition, it is impossible that there are other missing points, say $p$, between $p_{2}$ and $p_{3}$, that $\angle p > \epsilon$.

\item If more than one consecutive key-points are missing, the above method will only detect one missing point as an representation of all the missing key-points. For example, Fig.\ref{fig:key-point}(c) shows $p_{k12}$ and $p_{k22}$ are two missing key-points, however, one virtual key-point $p_{2}$ based on the existing points $p_{1}$, $p_{k11}$, $p_{k21}$, and $p_{3}$ are deduced.
\end{itemize}

Key-points capture the critical information and fill the missing information of a $PTS$, hence, the detected key-points can be used to represent the raw $PTS$. In the sequel sections, a $PTS$ typically refers to a series of key-points of the original $PTS$.

\subsection{Anomaly Detection in Corrected Time Series}
Anomaly detection and normal pattern identification are both processed based on the unit of $period$. The first step is to identify period points $Q$ that separate $PTS$ into a set of periods. We use clustering method to categorize the inflexions of a $PTS$ into a number of clusters. Then a cluster quality validation mechanism is applied to validate the quality of each cluster. The cluster with the highest quality will be adopted as the period cluster, that is, the points in the period cluster will be the period points for the time series. The period points are the points that can regularly and consistently separate the $PTS$ better than the points in the other clusters.

The cluster quality validation mechanism is a silhouette-value based method, in which the cluster that have highest mean silhouette value will be assumed to have the best clustering pattern. To accurately conduct clustering, we introduce a feature vector for each inflexion of $PTS$, with the optimal intention that each point can be distinguished with others efficiently.

\subsubsection{Time Series Compression: TSCC}
To save the storage space and improve the calculation efficiency, the raw $PTS$ will first be compressed. In this work, we use the Douglas–Peucker (DP) \cite{Hershberger1994} algorithm to compress a $PTS$, which is defined as: (1) use line segment $\overline{p_{1}p_{n}}$ to simplify the $PTS$; (2) find the farthest point $p_{f}$ from $\overline{p_{1}p_{n}}$; (3) if distance $d(p_{f},\overline{p_{1}p_{n}}) \leq \lambda$, where $ \lambda $ is a small value, and $ \lambda \geq 0 $, then the $PTS$ can be simplified by $\overline{p_{1}p_{n}}$, and this procedure is stopped; (4) otherwise, recursively simplify the subsequences $\{ p_{1}, \cdots, p_{f}\}$ and $\{ p_{f}, \cdots, p_{n}\}$ using steps $(1-3)$.

\begin{definition}
Given a $PTS = (v_{1},\cdots,v_{n})$, a \textbf{compressed time series} $CTS$ of $PTS$ is represented as $CTS = (v_{c_{1}},\cdots, v_{c_{n}}) \subseteq PTS$, where $\forall p_{c_{i}}\in CTS$ is an inflexion, and $|CTS|\ll |PTS|$.
\end{definition}
\begin{figure}
\centering
\includegraphics[width=7cm, height=4cm]{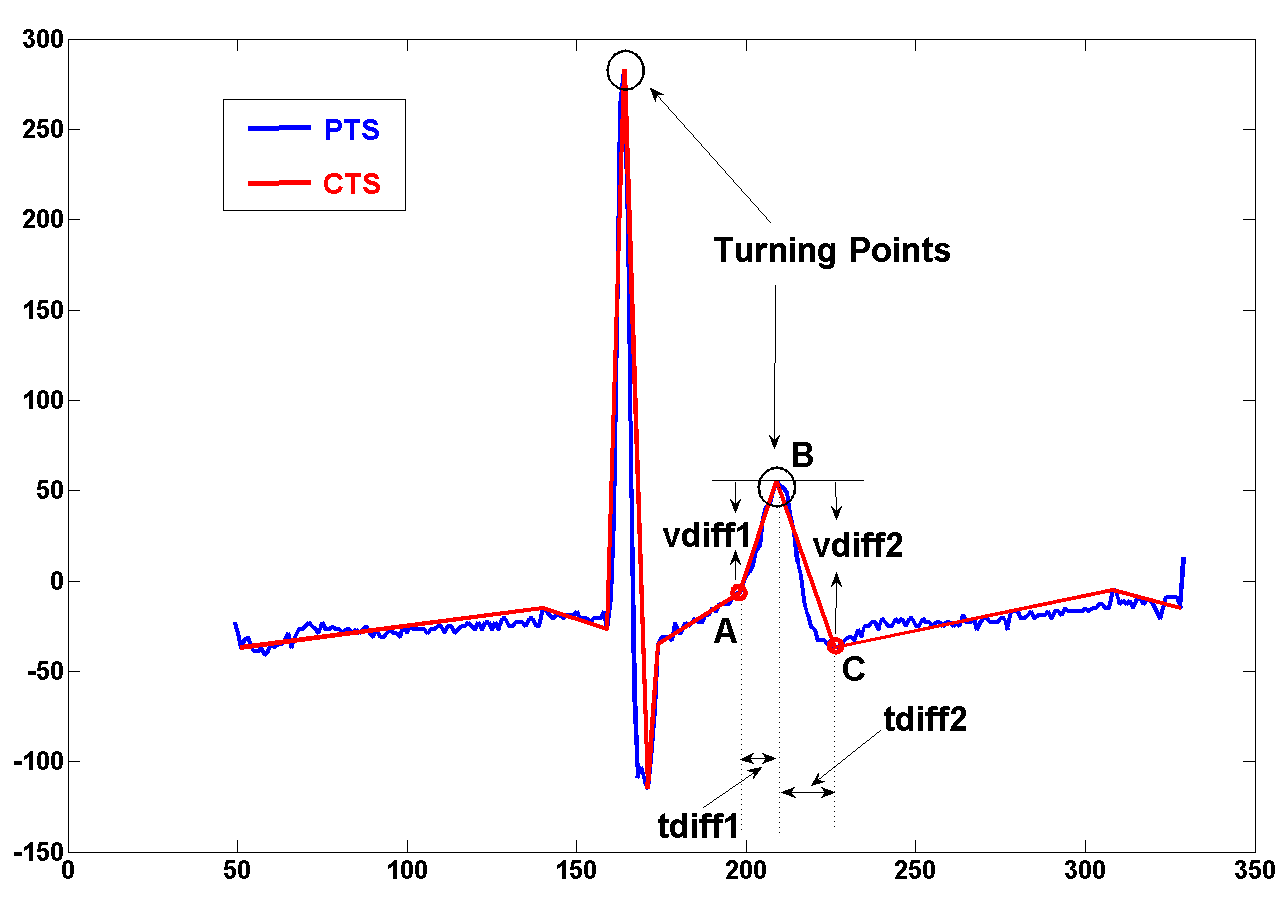}
\caption{A $PTS$ and one of its $CTSs$}
\label{fig:CTS}
\end{figure}

The feature vector of an inflexion is defined as:
\begin{definition}
A \textbf{feature vector} for a point $p_{i}\in PTS$ is a four-value vector $vec_{i} = (vdiff1_{i}, vdiff2_{i}, tdiff1_{i}, tdiff2_{i})$, where $vdiff1_{i} = v_{i} - v_{i-1}$, $vdiff2_{i} = v_{i+1} - v_{i}$, $tdiff1_{i} = t_{i} - t_{i-1}$, and $tdiff2_{i} = t_{i+1} - t_{i}$.
\end{definition}

\begin{example}
Fig.\ref{fig:CTS} shows an example of a $PTS$ and one of its compressed time series $CTS$. The value differences $vdiff1$ and $vdiff2$, and the time differences $wdiff1$ and $wdiff2$ are shown in Fig.\ref{fig:CTS}.
\end{example}

\subsubsection{Period Segmentation and Summarization: PSSC}
PSSC component identifies period points that separate the $CTS$ into a series of periods, which is implemented by three steps: cluster points of $CTS$, evaluate the quality of clusters based on silhouette value, and Segment and annotate periods. Details of these steps are given below.

\textbf{Step 1: Cluster Points of CTS} Points are clustered into a number of clusters based on their feature vectors. In this work, we use $k$-means++ \cite{arthur2007k} clustering method to cluster points. It has been validated that based on the proposed feature vector, the $k$-means++ is more accurate and less time-consumed than other clustering tools (e.g., $k$-means \cite{hartigan1979algorithm}, Gaussian mixture models \cite{reynolds2009gaussian} and spectral clustering \cite{ng2002spectral}). We give an brief introduction of the $k$-means++ in this section.

$k$-means++ is an improvement of $k$-means by first determining the initial clustering centres before conducting the $k$-means iteration process. $k$-means is a classical $NP$-hard clustering method. One of its drawbacks is the low clustering accuracy caused by randomly choosing the $k$ starting points. The arbitrarily chosen initial clusters cannot guarantee a result converging to the global optimum all the time. $k$-means++ is proposed to solve this problem. K-mean++ chooses its first cluster center randomly, and each of the remaining ones is selected according to the probability of the point's squared distance to its closest centre point being proportional to the squared distances of the other points. The $k$-means++ algorithm has been proved to have a time complexity of $O(logk)$ and it is of high time efficiency by determining the initial seeding. For more details of $k$-means++, readers can refer to \cite{arthur2007k}.

\textbf{Step 2: Evaluate the quality of clusters based on silhouette value}. We use the mean Silhouette value\cite{rousseeuw1987silhouettes} of a cluster to evaluate the quality of a cluster. The silhouette value can interpret the overall efficiency of the applied clustering method and the quality of each cluster such as the tightness of a cluster and the similarity of the elements in a cluster. The silhouette value of a point belonging to a cluster is defined as:

\begin{definition}
Let points in $PTS$ be clustered into $k$ clusters: $C_{CTS} = \{C_{1},\cdots,C_{m},\cdots,C_{k}\}, k \leq |CTS|$. For any point $p_{i} = v_{i} \in C_{m}$, the silhouette value of $p_{i}$ is
\begin{equation}
 sil(p_{i}) = \frac{b(p_{i})-a(p_{i})}{max\{a(p_{i}),b(p_{i})\}}
\end{equation}
where $a(p_{i}) = \frac{1}{M-1}\sum_{p_{i},p_{j} \in C_{m},i\neq j}sim(p_{i},p_{j}), M = |C_{m}|$ is the number of elements in cluster $m$; $b(p_{i}) = min(\frac{1}{M-1}\sum_{p_{i} \in C_{m},p_{j} \in C_{h},h\neq m}sim(p_{i},p_{j}))$. $sim(p_{i},p_{j})$ represents the similarity between $p_{i}$ and $p_{j}$.
\end{definition}

In the above definition, $sim(p_{i},p_{j})$ can be calculated by any similarity calculation formula. In this work, we adopt the Euclidean Distance as similarity measure, i.e., $sim(p_{i},p_{j}) = \sqrt{(v_{i}-v_{j})^{2}+(t_{i}-t_{j})^{2}}$, where $t_{i}$ and $t_{j}$ are the time indexes of the points $p_{i}$ and $p_{j}$. From the definition, $a(p_{i})$ measures the dissimilarity degree between point $p_{i}$ and the points in the same cluster, while $b(p_{i})$ refers to the dissimilarity between $p_{i}$ and the points in the other clusters. Therefore, a small $a(p_{i})$ and a large $b(p_{i})$ indicate a good clustering. As $-1 \leq sil(p_{i})\leq 1$, a $sil(p_{i}) \rightarrow 1$ means that a point $p_{i}$ is well clustered, while $sil(p_{i}) \rightarrow_{+} 0$ represents the point is close to the boundary between clusters $M$ and $H$, and $sil(p_{i}) < 0$ indicates that point $p_{i}$ is close to the points in the neighbouring clusters rather than the points in cluster $M$.

The mean value of the silhouette values of points is used to evaluate the quality of the overall clustering result: $msil(C_{CTS}) = \frac{1}{|CTS|}\sum_{p_{i}\in CTS} sil(p_{i})$. Similar to the silhouette value of a point, the $msil \rightarrow 1$ represents a better clustering.

\begin{algorithm}[t] \SetAlgoNoLine
\KwIn{$\;$ (1)$\; V=\{vec_{i}| 1 \leq i \leq |CTS| \}$, where $vec_{i} = (\alpha^{i},diff1_{i},diff2_{i})$ \\
$\qquad\quad\quad$(2)$\;$A set of point clusters: $C_{CTS} = \{C_{m}|1 \leq m \leq k\}$ \\
$\qquad\quad\quad$(3)$\;$Threshold values $\eta$ and $\xi$, $0 \leq \eta, \xi \leq 1$}
\KwOut{Period cluster $C_{perid}$}   Calculate $sil(p_{i})$ for $\forall p_{i} \in CTS$; \\   Calculate mean silhouette value: $msil(C_{CTS})$; \\ \If{$msil(C_{CTS}) < \eta $}{   $C_{perid}= NULL$; \\    return; \\ } $C_{perid} = max(msil(C_{m})) \; \& \; msil(C_{m}) > \xi$ for $\forall C_{m}\in C_{CTS}$. \caption{Cluster quality validation} \label{alg:clusterQualityValidation} \end{algorithm}

After clustering, we need to choose a cluster in which the points will be used as period points for the $CTS$. The chosen cluster is called \textbf{period cluster}. The points in the period cluster are the most stable points that can regularly and consistently separate $CTS$. We use the mean silhouette value of each cluster to evaluate the efficiency of a single cluster, represented as $msil(C_{m}) = \sum_{p_{i}\in C_{m}} sil(p_{i})$, where $-1 \leq msil(C_{m}) \leq 1$, and $msil(C_{m}) \rightarrow 1$ means the high quality of the cluster $m$. Based on the definition of silhouette values, we give \textbf{Algorithm \ref{alg:clusterQualityValidation}} of choosing period cluster from a clustering result. Algorithm \ref{alg:clusterQualityValidation} shows that if the mean silhouette value of the overall clustering result is less than a pre-defined threshold value $\eta$, then the clustering result is unqualified. Feature vectors of points need to be re-clustered with adjusted parameters, e.g., change the number of clusters. The last line indicates that the chosen period cluster is the one with highest mean silhouette values that is higher than a threshold $\xi$.

\textbf{Step 3.} Segmentation and annotation of periods. As mentioned in the previous section, a $CTS$ can be divided into a series of periods by using the period points. Thus detecting a local anomaly in $CTS$ means to identify an abnormal period or periods. In this section, we introduce a segmenting approach to extract the main and common features of each period. The extracted information will be used as classification features that are used for model learning and anomaly detection. In addition, signal annotations (e.g., '\textit{Normal}' and '\textit{Abnormal}') are attached to each period based on the original labels of the corresponding $PTS$. We will first give the concept of a summary of a period.
\begin{definition}
Given a CTS that has been separated into $D$ periods, a \textbf{summary} of a period $pd_{i} = (v_{i_{1}}, \cdots, v_{i_{m}}), 1 \leq i \leq D$ is a vector $seg_{i} = (h_{i}^{min},t_{i}^{min}, h_{i}^{max}, t_{i}^{max}, h_{i}^{mea},p_{i}^{minmax}, p_{i}^{l})$, where $h_{i}^{min}$ is the amplitude value of the point having minimum amplitude in period $i$: $h_{i}^{min} = min\{v_{i_{k}};1 \leq k \leq m\}$; $t_{i}^{min}$ is the time index of the point with minimum amplitude. If there are two points having the minimum amplitude, $t_{i}^{min}$ is the time index of the first point. $h_{i}^{max} = max\{v_{i_{k}}\}$; $t_{i}^{max}$ is the first point with maximum amplitude; $h_{i}^{mea} = \frac{1}{m}(\sum v_{i_{k}})$; $p_{i}^{minmax} = |t_{i}^{max} - t_{i}^{min}|$; $p_{i}^{l} = t_{i_{m}} - t_{i_{1}}$.
\end{definition}

We represent the segmented $CTS$ as $STS = \{seg_{1},\cdots,seg_{n}\}$. Each period corresponds to an annotation $ann$ indicating the state of the period. In this paper, we will only consider two states: $normal$ and $abnormal$. Therefore, a $STS$ is always associated with a series of annotations $A_{STS} = \{ann_{1},\cdots,ann_{n}\}$.

For the supervised pattern recognition model, the original $PTS$ has a set of labels to indicate the states of the disjoint sub-sequences of $PTS$, which are represented as $Lbs = \{lb_{(1)},\cdots, lb_{(w)}\}$, $\forall lb_{(r)} = \{'N'(Normal),'Ab'(Abnormal)\}, 1 \leq r \leq w$. However, $Lbs$ cannot be attached to the segmentations of the $PTS$ directly because the periodic separation is independent from the labelling process. To determine the state of a segmentation, we introduce a logical-multiplying relation of two signals:

\textbf{Rule 1.} \textit{$ann = \otimes('Ab','N')='Ab'$ and $ann = \otimes('N','N')='N'$}.

Assume a period covers a subsequence that is labelled by two signals, if there exists an abnormal behaviour in the subsequence, then based on rule 1, the behaviour of the segmentation of the period is abnormal; otherwise the period is a normal series. This label assignment rule can be extended to multiple labels: given a set of labels $Lbs= \{lb_{1},\cdots,lb_{r}\}$, if $\exists lb_{j} = 'Ab', 1 \leq j \leq r$, the value of $Lbs$ is $'Ab'$, represented as $lbs = \otimes(lb_{1},\cdots,lb_{r}) = 'Ab'$; if $\forall lb_{j} = 'N'$, $lbs = 'N'$.

According to the above discussion, the annotation of a period $pd_{i}$ is determined by \textbf{Algorithm \ref{alg:PeriodAnnotation}}.

 \begin{algorithm}[t]
 \SetAlgoNoLine
 \KwIn{Period $pd_{i} = (v_{i1},\cdots,v_{im}), 1 \leq i\leq n$\;
  A series of labels $Lbs = (lb_{1},\cdots,lb_{r})$\;
  }
 \KwOut{An annotated $pd_{i}^{'}$\;}
 $t_{i}^{1}=NULL$: the time of the $1^{st}$ annotation in the period\;
 $t_{i}^{end}=NULL$: the time of the last annotation in the period\;

 \If{$\exists lb_{j}$ that $t_{(i-1)1}\leq t_{j-1}\leq t_{(i-1)m} < t_{i1}\leq t_{j}\leq t_{im}$}
 {
 $t_{i}^{1} = t_{j}$\;
 }
 \If{$\exists lb_{k}$ \& $t_{i1} \leq t_{k} \leq t_{im}$ \& $t_{(i+1)1} \leq t_{k+1} \leq t_{(i+1)m}$}
 {
  $t_{i}^{end} = t_{k}$\;
 }

 \eIf{$t_{i}^{1} \neq NULL$ $\|$ $t_{i}^{end} \neq NULL$}
 {
 \If{$t_{i}^{1} = NULL$}
 {
  $t_{i}^{1} = 'N'$
 }
  \If{$t_{i}^{end} = NULL$}
  {
  $t_{i}^{end} = 'N'$
  }
  $Lbs = Lbs\{t_{i}^{1},\cdots,t_{i}^{end}\}$\;
  $lbs = \otimes(Lbs)$\;
 }
 {
 $lbs = Lbs\{t_{i+1}^{1}\}$\;
 }
\caption{Period annotation}
\label{alg:PeriodAnnotation}
\end{algorithm}

\begin{figure}
\centering
\includegraphics[width=8cm, height=4cm]{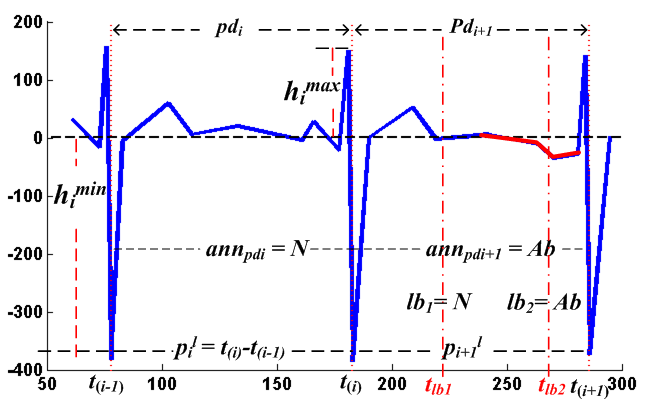}
\caption{Segmentation and annotation of two periods}
\label{fig:segann}
\end{figure}

\begin{example}
We present the segmentation and annotation of a period in Fig.\ref{fig:segann} to explain their processes more clearly. Fig.\ref{fig:segann} shows that $pd_{i}$ does not involve any label and the first label in $pd_{i+1}$ is $lb_{1} = N$, so $lb_{pd_{i}} = 'N'$. $lb_{2}$ is '$Ab$', hence $pd_{i+1}$ is annotated as '$Ab$'.
\end{example}

\subsubsection{Classification-based Anomaly Detection and Prediction: ADPC}
From Definition 4.6, each period of a $PTS$ is summarised by seven features of the period: $(h_{i}^{min},t_{i}^{min}, h_{i}^{max}, t_{i}^{max}, h_{i}^{mea},p_{i}^{minmax}, p_{i}^{l})$. Using these seven features to abstract a period can significantly reduce the computational complexity in a classification process. In the next section, we validate the proposed anomaly detection framework with various classification methods on the basis of different ECG datasets.

\section{Experimental Evaluation}
Our experiments are conducted in four steps. The first step is to compress the raw ECG time series by utilizing the DP algorithm, and to represent each inflexion in the perceived $CTS$ as a feature vector (see Definition 4.4). Secondly, the $K$-means++ clustering algorithm is applied to the series of feature vectors of the $CTS$, and the clustering result is validated by silhouette values. Based on the mean silhouette value of each cluster, a period cluster is chosen and the $CTS$ is periodically separated to a set of consistent segments. Thirdly, each segment is summarised by the seven features (see Definition 4.6). Finally, a normal pattern of the time series is constructed and anomalies are detected by utilizing classification tools on the basis of the seven features.

We validate the proposed framework on the basis of eight ECG datasets \cite{goldberger2000physiobank}, which are summarised in Table \ref{tab:datasets} where 'V' represents Premature ventricular contraction, 'A': Atrial premature ventricular, and 'S': Supraventricular premature beat. Apart from the $aftdb$ dataset, each time series is separated into a series of subsequences that are labelled by the dataset provider. We give the number of abnormal subsequences ('\#ofAbnor') and the number of normal subsequences ('\#ofNor') of each time series in Table \ref{tab:datasets}.

Our experiment is conducted on a $32$-bit Windows system, with $3.2$GHz CPU and $4$GB RAM. The ECG datasets are downloaded to a local machine using the WFDB toolbox \cite{HSilva2014,Goldberger2000} for $32$-bit MATLAB. We use the $10$-fold cross validation method to process the datasets.

\begin{table}
\tbl{ECG Datasets used in experiments\label{tab:datasets}}{
\begin{tabular}{l*{6}{c}r}
\hline
Datasets & Abbr. & \#ofSamples & AnomalyTypes & \#ofAbnor & \#ofNor \\
\hline
AHA0001 & ahadb & 899750 & V & 115 & 2162 \\
SupraventricularArrhythmia800 & svdb & 230400 & S \& V & 75 & 1846 \\
SuddenCardiacDeathHolter30 & sddb & 22099250 & V & 38 & 5743 \\
MIT-BIH Arrhythmia100 & mitdb & 650000 & A \& V & 164 & 2526 \\
MIT-BIH Arrhythmia106 & mitdb06 & 650000 & A \& V & 34 & 2239 \\
MGH/MF Waveform001 & mgh & 403560 & S \& V & 23 & 776 \\
MIT-BIH LongTerm14046 & ltdb & 10828800 & V & 000 & 000 \\
AF TerminationN04 & aftdb & 7680 & NA & NA & NA\\
\hline
\end{tabular}}
\end{table}

The metrics used for evaluating the final anomaly classification results include: \\
(1) Accuracy (acc): $(TP+TN)$ / Number of all classified samples; \\
(2) Sensitivity (sen): $TP$ / $(TP+FN)$; \\
(3) Specificity (spe): $TN$ / $(FP+TN)$; \\
(4) Prevalence (pre): $TP$ / Number of all samples. \\
(5) Fmeasure (fmea): $2*\frac{precision*recall}{precision+recall}$, where $recall = sen$, $precision = \frac{TP}{TP+FP}$ \\
$TP$ = true positive, $TN$ = true negative, $FP$ = false positive, and $FN$ = false negative.

Details of the experiments are illustrated in the following sections.
\subsection{Inflexion Detection and Time Series Compression}
At first, we design an experiment to detect the inflexions in a time series. The detected inflexions will be used as an approximation of the raw time series, and will be compressed by DP algorithm. We design this experiment based on the work of \cite{Rosin2003505}. We assess the stability of the uncertainty detection and DP compression algorithms under the variations of the change of scale parameters and the perturbation of data. The former is measured by using a monotonicity index and the latter is quantified by a break-point stability index.

The monotonicity index is used to measure the monotonically decreasing or increasing trend of the number of break points when the values of scale parameters of a polygonal approximation algorithm are changed. For the inflexion detection algorithm and the DP algorithm, if the values of the scale parameters $\epsilon$ and $\lambda$ are increasing, the number of the produced breakpoints of the time series will be decreasing, and vice versa. The decreasing monotonicity index is defined as $M_{D} = (1 - \frac{T_{+}}{T_{-}}) \times 100$, and the increasing monotonicity index is $M_{I} = (1 - \frac{T_{-}}{T_{+}}) \times 100$, where $T_{-} = - \sum_{ \forall \Delta v_{i} < 0} \Delta v_{i} / h_{i}$, $T_{+} = \sum_{ \forall \Delta v_{i} > 0} \Delta v_{i} / h_{i}$, and $h_{i} = \frac{v_{i} + v_{i-1}}{2}$. Both of $M_{D}$ and $M_{I}$ are in the range $[0,100]$, and their perfect scores are $100$.

\begin{table}
\tbl{Decreasing monotonicity degree of six datasets in terms of the value of $\epsilon$ and $\lambda$\label{tab:MD}}{
\begin{tabular}{l*{7}{c}r}
\hline
 & ahadb & svdb & sddb & mitdb & mgh & aftdb \\
\hline
$\epsilon$ & 100 & 100 & 100 & 100 & 100 & 100 \\
$\lambda$ & 100 & 100 & 100 & 100 & 100 & 100 \\
\hline
\end{tabular}}
\end{table}

\begin{figure}
\centering
\includegraphics[width=12cm, height=5cm]{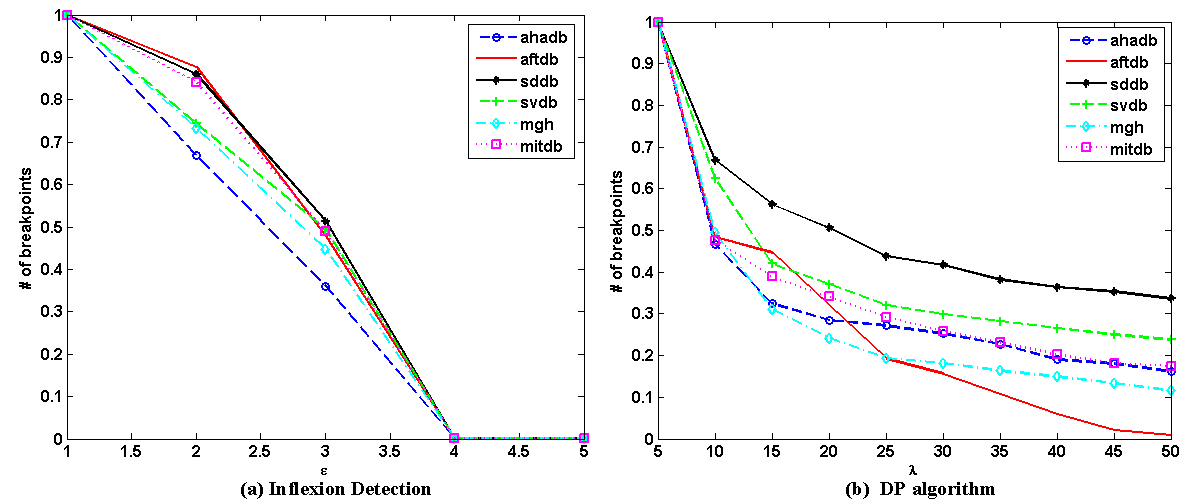}
\caption{Monotonically decreasing number of breakpoints in terms of $\epsilon$ for the inflexion detection procedure and $\lambda$ for the DP algorithm}
\label{fig:MD}
\end{figure}

We test the decreasing monotonicity degrees for the datasets $ahadb$, $svdb$, $sddb$, $mitdb$, $mgh$, and $aftdb$ in terms of different values of $\epsilon$ for inflexion detection procedure and $\lambda$ for DP algorithm. For the inflexion detection procedure, we set $\epsilon = 1, 2, 3, 4, 5$. Table \ref{tab:MD} shows that the breakpoint numbers for the six datasets are perfectly decreasing in terms of the increasing $\epsilon$, which can also be seen in Fig.\ref{fig:MD}(a). For DP algorithm, we first fix $\epsilon = 1$, and detect inflexions of the six time series. Based on the detected inflexions, we set $\lambda = 5, 10, 15, 20, 25, 30, 35, 40, 45, 50$ to conduct DP compression. From Table \ref{tab:MD} and Fig.\ref{fig:MD}(b), we can see that the numbers of breakpoints are also $100\%$ decreasing in terms of the increasing $\lambda$.

The break-point stability index is defined as the shifting degree of breakpoints when deleting increasing amounts from the beginning of a time series. We use the endpoint stability to test the breakpoint stability for fixed parameter settings : $\epsilon = 1$ for the inflexion detection and $\lambda = 10$ for the DP algorithm. The endpoint stability measurement is defined as $S = (1-\frac{1}{m}\sum_{d}\sum_{b}\frac{s_{b}^{d}}{n_{d}l_{d}})$, where $m$ is the level number of deletion, $d$ is the $dth$ level, $s_{b}^{d}$ is the shifting pixels at breakpoint $b$, $l_{d}$ is the length of the remaining time series and $n_{d}$ is the number of breakpoints after the $dth$ deletion. Table \ref{tab:pertub} shows the deletion length of each running circle and the stability degree of each time series. For example, after inflexion detection, the sample number of $ahadb$ is $307350$. We iteratively delete $10000$ samples from the beginning of the remaining $ahadb$ time series, and conduct the DP algorithm based on the new time series. The positions of the identified breakpoints in each running circle are compared with the positions of the breakpoints identified in the whole $ahadb$. From Table \ref{tab:pertub}, we can see that each time series is of high stability (i.e. values of $S$) when conducting the uncertainty detection procedure and the DP algorithm with fixed scale parameters.

\begin{table}
\tbl{Endpoint stability of six datasets and pertubations\label{tab:pertub}}{
\begin{tabular}{l*{7}{c}r}
\hline
 & ahadb & svdb & sddb & mitdb & mgh & aftdb \\
\hline
Shifting length & 10000 & 10000 & 10000 & 10000 & 10000 & 100 \\
$S$ & 100 & 99.8988 & 99.9955 & 99.9725 & 99.9348 & 99.9351 \\
\hline
\end{tabular}}
\end{table}

\subsection{Compressed Time Series Representation}

\begin{figure}
\centering
\includegraphics[width=12cm, height=9cm]{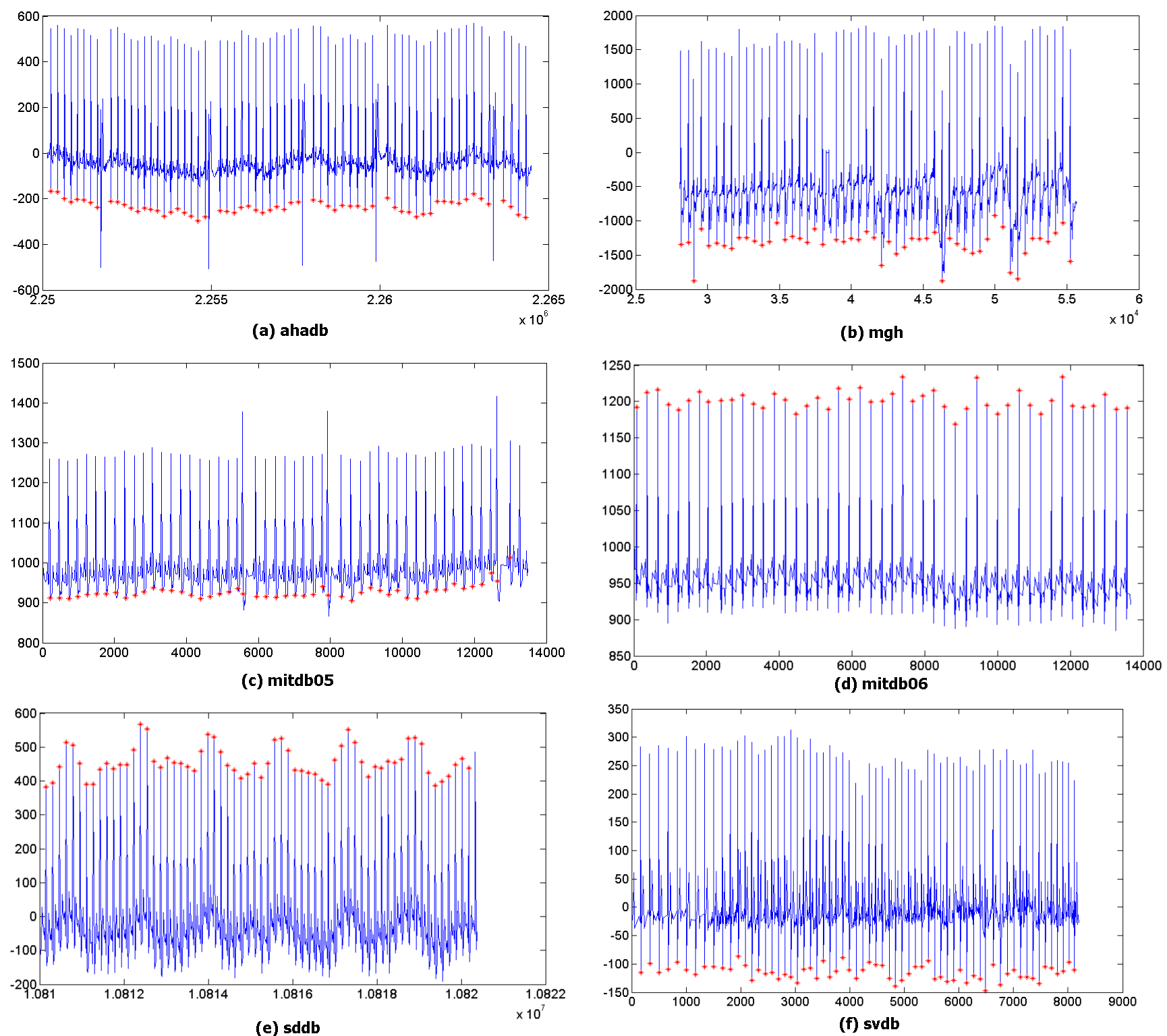}
\caption{Period point identification of four datasets based on feature vectors}
\label{fig:pdps}
\end{figure}

From the above testing (see Fig.\ref{fig:MD}), we can see that when $\epsilon \ge 4$, the number of detected inflexions of each time series is going to be $0$. Based on Fig.\ref{fig:MD}, we set $\epsilon = 1$ and $\lambda = 10$ for inflexion detection and time series compression. We then compare three methods of period point representation: (1) inflexions in $CTS$ are represented by feature vectors (FV); (2) inflexions are represented by angles (Angle) of peak points \cite{huang2013online}; (3) inflexions are represented by valley points (Valley) \cite{tang2007effective}. Valley points are points in a $PTS$, which have values less than an upper bound value (represented as $U$). $U$ is initially specified by users and will be updated as time evolves. The update procedure is defined as $U_{b} = \alpha(\sum_{i=1}^{N}V_{i})/N$, where $N$ is the number of past valley points and $\alpha$ is an outlier control factor that is determined and adjusted by experts. As stated by Tang et al.\cite{tang2007effective}, the best values of initial upper bound and $\alpha$ in ECG are $50mmHg$ and $1.1$. The perceived feature vector sets, angle sets, and valley point sets are passed to the next step in which points are clustered and the period points of the $CTS$ are identified. Each period is then segmented using the proposed segmentation method(see Definition 4.6). Finally, Linear Discriminant Analysis (LDA) and Naive Bayes(NB) classifiers are applied for sample classification and anomaly detection. Fig \ref{fig:pdps} shows the identified period points using the FV-based method for four datasets: $ltdb$, $sddb$, $svdb$ and $ahadb$. From Fig \ref{fig:pdps}, we can see that for each dataset, the FV-based method successfully identifies a set of periodic points that can separate the $CTS$ in a stable and consistent manner.

Table \ref{tab:silhouette} presents the silhouette values of clustering the inflexions in the $CTSs$ of seven time series, where column 'mean' refers to the mean silhouette value of a dataset clustering, and the values in columns c(luster)1-6 are the mean silhouette values of each cluster after clustering a dataset. 'NAs' in the sixth column means that the inflexions in the corresponding datasets are clustered into five groups, which present the best clustering performance in this dataset. From Definition 4.5, we know that if the silhouette values in a cluster is close to $1$, the cluster includes a set of points having similar patterns. On the other hand, if the silhouette values in a cluster are significantly different from each other or have negative values, the points in the cluster have very different patterns with each other or they are more close to the points in other clusters. Table \ref{tab:silhouette} shows that for each of the seven datasets, the mean silhouette values of the overall clustering result and each of the individual clusters are higher than $0.4$ ($\eta = 0.4$ in algorithm \ref{alg:clusterQualityValidation}). The best silhouette value of an individual cluster in each dataset is close to or higher than 0.9 ($\xi = 0.8$ in Algorithm \ref{alg:clusterQualityValidation}). In addition, for each dataset, we select the points in the cluster with highest silhouette value as the period points. For example, for dataset $ahadb$, points in cluster $4$ are selected as period points.
\begin{table}
\centering
\tbl{Silhouette values of six datasets\label{tab:silhouette}}{
\begin{tabular}{ c c c c c c c c }
 \hline
 Dataset & \multicolumn{7}{c}{Silhouette values} \\
 \hline
 & mean & cluster1 (c1) & c2 & c3 & c4 & c5 & c6 \\
 \hline
 ahadb & 0.8253 & 0.4479 & 0.8502 & 0.9824 & 0.9891 & 0.9381 & NA \\
 svdb & 0.6941 & 0.9792 & 0.6551 & 0.9703 & 0.5463 & 0.5729 & 0.959 \\
 sddb & 0.772 & 0.6888 & 0.5787 & 0.965 & 0.9727 & 0.6971 & 0.7529 \\
 mitdb & 0.9373 & 0.9877 & 0.7442 & 0.9898 & 0.9711 & 0.5854 & 0.3754 \\
 mitdb06 & 0.7339 & 0.7317 & 0.8998 & 0.609 & 0.8577 & 0.8669 & NA \\
 ltdb & 0.9149 & 0.9164 & 0.8381 & 0.9739 & 0.9079 & 0.8975 & NA \\
 mgh & 0.8253 & 0.4479 & 0.8502 & 0.9824 & 0.9891 & 0.9381 & NA \\
 \hline
\end{tabular}}
\end{table}

\begin{figure}
\centering
\includegraphics[width=9cm, height=5cm]{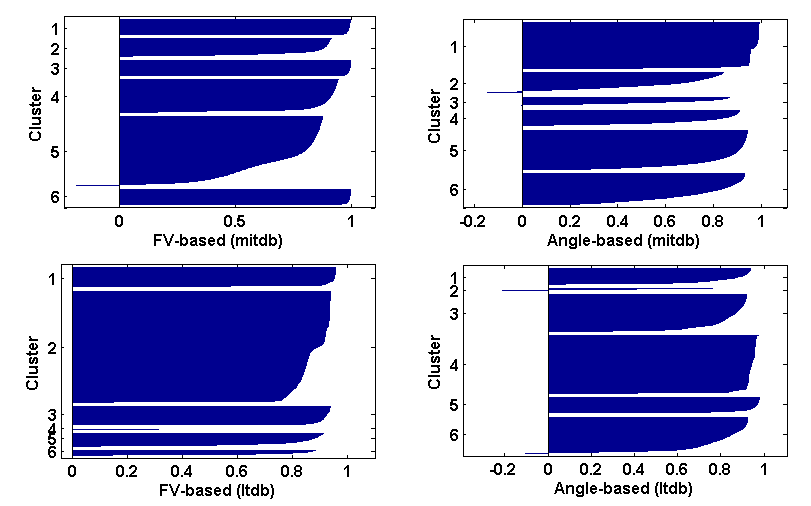}
\caption{Silhouette value comparison between the feature vector based clustering method (FV-based) and the angle-based clustering method for the $mitdb$ and $ltdb$ datasets}
\label{fig:sil_mitdb_ltdb}
\end{figure}

Fig \ref{fig:sil_mitdb_ltdb} presents the silhouette values of clustering the inflexions in the $CTSs$ of $mitdb$ and $ltdb$ time series. From this figure, we can see that for both the $mitdb$ and $ltdb$ datasets, FV-based clustering results in fewer negative silhouette values in all clusters, and he values in each cluster are more similar to each other compared with the angle-based clustering. We also come to a similar conclusion by examining their mean silhouette values. The mean silhouette values of FV-based clustering for $mitdb$ (corresponding to Fig.\ref{fig:sil_mitdb_ltdb}(a)) is $0.9373$, while the angle-based clustering (Fig.\ref{fig:sil_mitdb_ltdb}(b)) is $0.7461$; and the mean values for $ltdb$ are $0.9149$ and $0.8155$ (Fig.\ref{fig:sil_mitdb_ltdb}(c) and Fig.\ref{fig:sil_mitdb_ltdb}(d)) respectively.

\begin{figure}
\centering
\includegraphics[width=10cm, height=5cm]{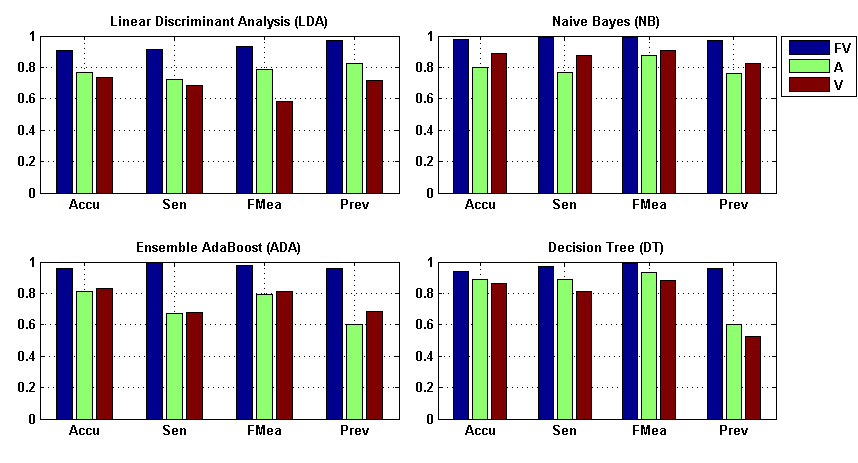}
\caption{Average performance comparison of four classifiers (LDA, NB, ADA, DT) based on feature vector based (FV), angle based (A) and valley point based (V) periodic separating methods}
\label{fig:Perf3}
\end{figure}

Fig.\ref{fig:Perf3} compares the average classification performance on the basis of four datasets using four classifiers: LDA, NB, Decision tree (DT), and AdaBoost (Ada) with $100$ ensemble members. From Fig.\ref{fig:Perf3}, we can see that the classifiers based on the FV periodic separating method have the best performance in terms of the four datasets (i.e., the highest accuracy, sensitivity, f-measure, and prevalence). In the case of LDA and DT, the valley-based periodic separating method has the worst performance while in the cases of NB and Ada, valley-based methods perform better than angle-based methods.

\subsection{Evaluation of Classification Based on Summarized Features}

This section describes the experimental design and the performance evaluation of classification based on the summarized features. This experiment is conducted on seven datasets: $ahadb$, $svdb$, $sddb$, $mitdb$, $mitdb06$, $mgh$, and $ltdb$. From the previous subsections, we know that the seven time series have been compressed and the period segmenting points have been identified (see Table \ref{tab:silhouette}). The segments of each of the time series are classified by using three classification tools: Random Forest with 100 trees (RF), LDA and NB. We use matrices of $acc$, $sen$, $spe$, and $pre$ to validate the classification performance.

\begin{figure}
\centering
\includegraphics[width=10cm, height=7cm]{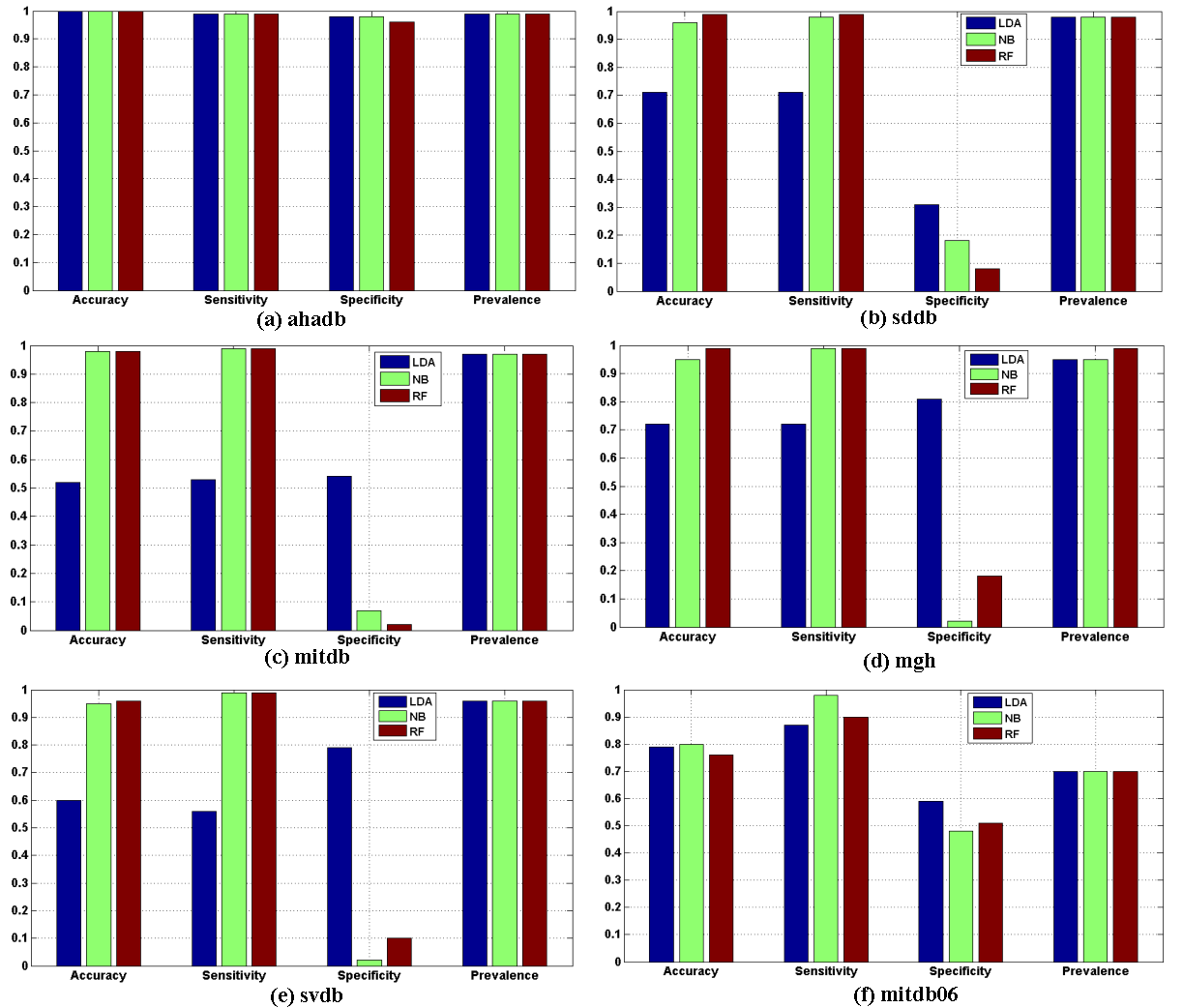}
\caption{Classification performance of six datasets based on the summarized features using classification methods of LDA, RF, and NB}
\label{fig:SixPerformanceCombine}
\end{figure}

The classification performance is shown in Fig.\ref{fig:SixPerformanceCombine}, which compares the performance of classification methods LDA, NB and RF, based on datasets (a) \textit{ahadb}, (b) \textit{sddb}, (c) \textit{mitdb}, (d) \textit{mgh}, (e) \textit{svdb}, and (f) \textit{mitdb06}. From the figure, we can see that for all six datasets, the performances of NB and RF are better than the performance of LDA based on the selected features. The accuracy and sensitivity of NB and RF are higher than $80\%$ for each of the datasets. Their prevalence values are over $90\%$ for the first five datasets (a-e). However, we can also see that the feature values of LDA are always higher than the feature values of the other two methods.

\begin{figure}
\centering
\includegraphics[width=14cm, height=6cm]{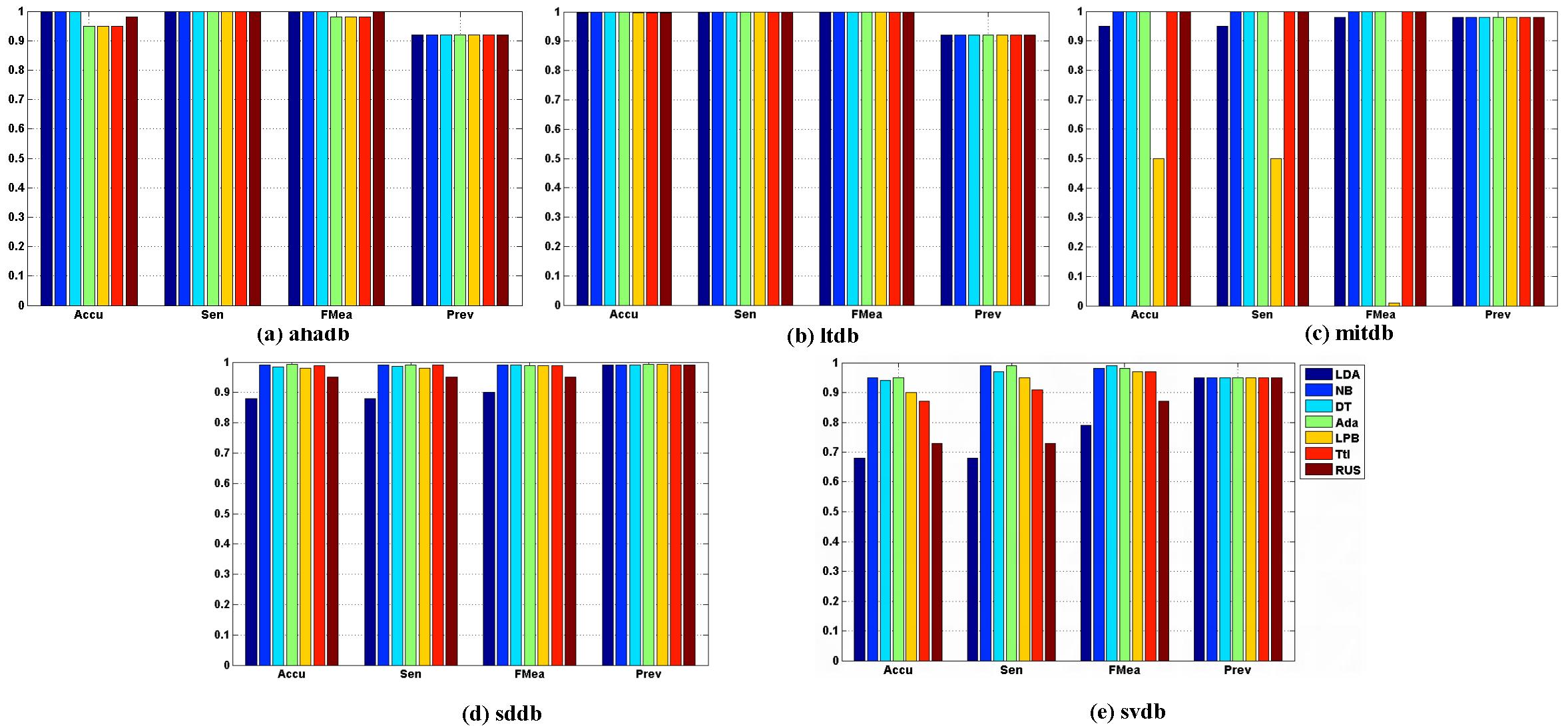}
\caption{Performance of seven classifiers (LDA, NB, DT, Ada, LPB, Ttl, and RUS) based on the proposed period identification and segmentation methods on five datasets ((a) ahadb, (b) ltdb, (c) mitdb, (d) sddb, and (e) svdb)}
\label{fig:PerfClassBarPlot5}
\end{figure}

\subsection{Performance Evaluation of Other Classification Methods Based on Summarized Features}

 In this section, we design an experiment to evaluate the performance of the proposed time series segmentation method. Experimental results on the basis of five datasets (i.e.,$mitdb$, $ltdb$, $ahadb$, $sddb$ and $svdb$) are presented in this section. We carry out the experiment by the following steps. First, the raw time series are compressed by DP algorithm and periodically separated by feature vector based period identification method. Second, each period is summarized by the proposed period summary method (see Definition 4.7) and is annotated by the annotation process(see Section 4.3). The classification methods used in this experiment include LDA, NB, DT, and a set of ensemble methods: AdaBoost (Ada), LPBoost (LPB), TotalBoost (Ttl), and RUSBoost (RUS). The classification performance is validated by five benchmarks: $acc$, $sen$, $fmea$, and $prev$.

Fig.\ref{fig:PerfClassBarPlot5} shows the evaluated results of the classifier performance based on the proposed period identification and segmentation method. From Fig.\ref{fig:PerfClassBarPlot5}, we can see that the accuracy values of classification based on the $5$ datasets are over $90\%$, except the cases of LPB with $mitdb$, LDA with $sddb$, LDA with $svdb$, and RUS with $svdb$. Some of them are of more than $98\%$ accuracy. The sensitivity of classification based on the datasets of $ahadb$, $ltdb$, and $mitdb$ are closing to $100\%$. The sensitivity based on the datasets of $sddb$ and $svdb$ are over $85\%$. The f-measure rates of classification based on $ahadb$, $ltdb$, $mitdb$, and $sddb$ are higher than $95\%$. The f-measure rates of RUS and LDA based on $mitdb$ and $svdb$ are less than $80\%$, but the f-measure of other classifiers based on these two datasets are all higher than $80\%$, and some of them are closing to $100\%$. The prevalence rates of classification on the basis of the five datasets are over $90\%$.

\section{Conclusions}
In this paper, we have introduced a framework of detecting anomalies in uncertain pseudo periodic time series. We formally define pseudo periodic time series ($PTS$) and identified three types of anomalies that may occur in a $PTS$. We focused on local anomaly detection in $PTS$ by using classification tools. The uncertainties in a $PTS$ are pre-processed by an inflexion detecting procedure. By conducting DP-based time series compression and feature summarization of each segment, the proposed approach significantly improves the time efficiency of time series processing and reduces the storage space of the data streams. One problem of the proposed framework is that the silhouette coefficient based clustering evaluation is a time consuming process. Though the compressed time series contains much fewer data points than the raw time series, it is necessary to develop a more efficient evaluation approach to find the optimal clusters of data stream inflexions. In the future, we are going to find a more time efficient way to recognize the patterns of a $PTS$. In addition, we will do more testing based on other datasets to further validate the performance of the method. Correcting false-detected inflexions and detecting global anomalies in an uncertain $PTS$ will be the main target of our next research work.

\section*{Acknowledgement}
This work is supported by the National Natural Science Foundation of China (NSFC 61332013) and the Australian Research Council (ARC) Discovery Projects DP140100841, DP130101327, and Linkage Project LP100200682.

\bibliographystyle{ACM-Reference-Format-Journals}
\bibliography{bibDM}

\end{document}